\documentclass[10pt,twocolumn,letterpaper]{article}

\usepackage[pagenumbers]{cvpr} 

\usepackage{graphicx}
\usepackage{amsmath}
\usepackage{amssymb}
\usepackage{booktabs}
\usepackage{bbm}

\usepackage[pagebackref,breaklinks,colorlinks]{hyperref}

\usepackage[capitalize]{cleveref}
\crefname{section}{Sec.}{Secs.}
\Crefname{section}{Section}{Sections}
\Crefname{table}{Table}{Tables}
\crefname{table}{Tab.}{Tabs.}
\Crefname{eqn}{Eqn.}{Eqs.}

\def\cvprPaperID{*****} 
\def\confName{CVPR}
\def\confYear{2022}

\newcommand\norm[1]{\left\lVert#1\right\rVert}

\begin{document}

\title{Overfitting In Contrastive Learning?}

\author{Zachary Rabin, Jim Davis\\
Ohio State University\\
{\tt\small \{rabin.30, davis.1719\}@osu.edu}
\and
Benjamin Lewis, Matthew Scherreik\\
Air Force Research Laboratory\\
{\tt\small \{benjamin.lewis.13, matthew.sherreik.1\}@us.af.mil}
}
\maketitle

\begin{abstract}
    Overfitting describes a machine learning phenomenon where 
    the model fits too closely to the training data, resulting in poor generalization. 
    While this occurrence is thoroughly documented for many forms of supervised learning, it is not well examined in the context of \underline{un}supervised learning.
    In this work we examine the nature of overfitting in  unsupervised contrastive learning. 
    We show that overfitting can indeed occur and the mechanism behind overfitting.
\end{abstract}

\section{Introduction}
\label{sec:intro}

Overfitting in neural networks is commonly observed. 
When selecting how long to train a network, a simple choice is to train until overfitting occurs. 
One can detect overfitting in the classical supervised setting by observing that the training error decreases, while validation error begins to increase and diverge from the training error. 
This signals that the network has ``overfit'' to the training data. 
We ask if overfitting can occur with unsupervised contrastive learning, and if so, what does overfitting mean in this context.

\section{Related Work}
\label{sec:related}

Contrastive learning is a paradigm exploiting the similarities between positive and negative samples to learn features. 
While this paradigm can be supervised or unsupervised, SimCLR \cite{SimCLR} is a popular unsupervised contrastive learning approach, which we use in this work.

The SimCLR architecture consists of a backbone network $f(\cdot)$ and a projection head $g(\cdot)$. For some image $x$, we define $h = f(x)$, and $z=g(h) = g(f(x))$ 

Given a minibatch of size $N$, SimCLR first creates two augmented views of each source image.
We denote the pair of images created from the same source image as positive pairs.
Pairs of images created from different source images are referred to as negative pairs.
The $2N$ augmented images are arranged so that positive pairs appear consecutively.
Then, the loss function for a positive pair of images $(x_i,x_j)$ is defined as

\begin{equation}
    \ell(x_i,x_j) = -\text{log} \frac{\text{exp} ( \text{sim} (z_i,z_j)/\tau) } 
    {\sum_{k=1}^{2N}\mathbbm{1}_{[k\neq i]} \text{exp} ( \text{sim} (z_i,z_k)/\tau)}
    \label{eq:simclr}
\end{equation}


\noindent where $\tau$ is a temperature hyperparameter and sim($u,v$) = $u^\top v / \norm{u} \norm{v}$. The overall loss for the batch is

\begin{equation}
    \mathcal{L} = \frac{1}{2N}\sum_{k=1}^{N} [\ell(x_{2k-1}, x_{2k}) + \ell(x_{2k}, x_{2k-1})]
    \label{eq:simclr2}
\end{equation}

\noindent the average of the loss for each positive pair. 

\section{Method}

Rearranging \cref{eq:simclr} yields 


\begin{equation}
\begin{aligned}[b]
    \ell(x_i,x_j) &=-\text{sim}(z_i,z_j)/\tau + \\
    &\text{log} \sum_{k=1}^{2N}\mathbbm{1}_{[k\neq i]} \text{exp} ( \text{sim} (z_i,z_k)/\tau)
\end{aligned}
\end{equation}

\noindent This reveals a two part loss function where the first term encourages positive pairs to be similar to each other, while the second term encourages negative pairs to be farther apart. 
For convenience we call $-\text{sim}(z_i,z_j)/\tau$ the positive similarity and $\text{log} \sum_{k=1}^{2N}\mathbbm{1}_{[k\neq i]} \text{exp} ( \text{sim} (z_i,z_k)/\tau)$ the negative similarity.

Our experiments consist of running the SimCLR approach for up to 5K epochs to observe if overfitting occurs.
For data shown in Figs \ref{fig:train_val} and \ref{fig:ab}, we use a subset of CIFAR10 \cite{cifar} consisting of only the ``Airplane'', ``Automobile'', ``Ship'', and ``Truck'' classes.
In our experiments we track the overall loss $\mathcal{L}$ as well as the average positive and negative similarity per epoch. 
We train a ResNet18 \cite{resnet} with the Adam \cite{Adam} optimizer with a learning rate of 0.001.
Our augmentation scheme is comprised of mean padding followed by random cropping, random horizontal flipping, random color jittering, grayscaling, and finally Gaussian blurring.

\section{Results and Discussion}

\begin{figure}
    \centering
    \includegraphics[width=0.95\columnwidth]{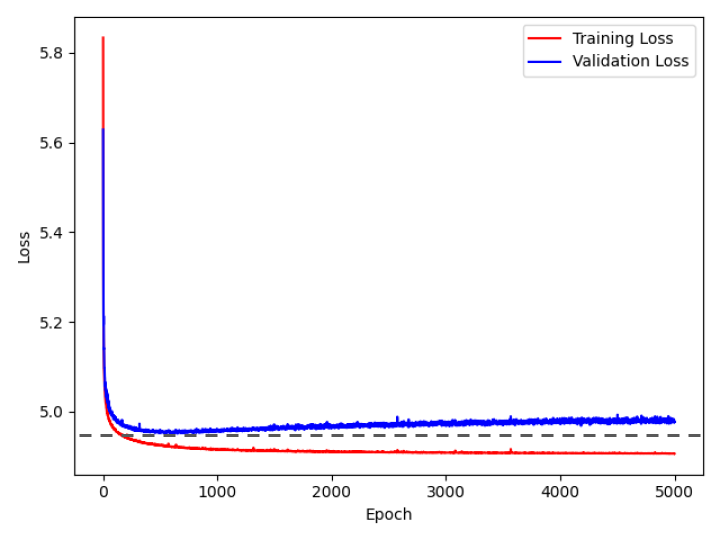}
    \caption{SimCLR overfitting on a subset of CIFAR10 over 5K epochs. A horizontal dashed line is provided as a reference.}
    \label{fig:train_val}
\end{figure}

\begin{figure}
    \centering
    \includegraphics[width=0.95\columnwidth]{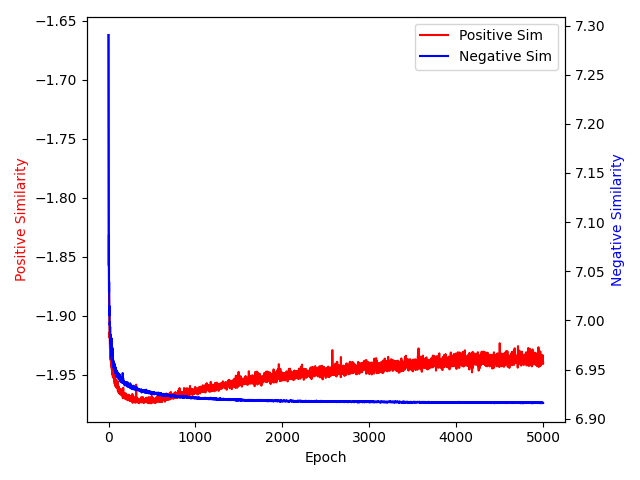}
    \caption{
    Validation loss of Fig. \ref{fig:train_val} split into positive and negative similarity.}
    \label{fig:ab}
\end{figure}

Firstly, we find that the loss $\mathcal{L}$ can in fact overfit if given enough time. 
We observe in Fig. \ref{fig:train_val} that the validation loss begins to diverge from the training loss. 
Using the dashed horizontal line as a reference we can see that the validation loss begins to increase over time after epoch 500. 

Secondly, we observe that the positive similarity is the term that drives the overfitting. 
Figure \ref{fig:ab} shows that the validation positive similarity initially decreases, but quickly begins to increase again, while the negative similarity continues to decrease. 
This shows that the positive similarity is the portion of the loss that drives the overfitting of the overall loss $\mathcal{L}$.
Conceptually, as training continues the model learns how to decrease the positive similarity for only training examples, and loses the capacity to detect positive pairs outside of the training set. 
This results in the representations $z$ for \textit{all} validation examples being pushed away from each other, even positive pairs. 


Understanding how contrastive learning overfits can be useful for reduced training times and achieving desired characteristics in the feature space.
As shown in Figs. \ref{fig:train_val} and \ref{fig:ab}, overfitting in the positive similarity can be detected earlier in the training process as opposed to overfitting in the overall loss. 
Therefore, stopping training when overfitting is seen in the positive similarity can reduce training times.

\section{Conclusion}

Unsupervised contrastive learning is a popular paradigm in deep-learning. 
In this work, we investigated if this framework is capable of overfitting and what overfitting means in this context. 
We found that given enough epochs, unsupervised contrastive learning can indeed overfit to the training set. 
We showed that when the model does overfit, it loses the ability to bring positive pair's representations closer to each other, resulting in the positive similarity being the driving factor behind the overfitting phenomenon. 

\section{Acknowledgements}

This work was supported in part by the U.S. Air Force
Research Laboratory under contract FA8650-21-C-1174.
Distribution A: Cleared for Public Release. 
Distribution Unlimited. PA Approval \#AFRL-2024-3668.

{\small
\bibliographystyle{ieee_fullname}
\bibliography{egbib}
}

\end{document}